\documentclass[10pt,conference,a4paper]{IEEEtran}

\usepackage{amsfonts}
\usepackage{bm}
\usepackage{subfigure}
\usepackage{pifont}
\usepackage{color}
\newcommand{\cmark}{\ding{51}}%
\newcommand{\xmark}{\ding{55}}%

\usepackage{cite}
\ifCLASSINFOpdf
   \usepackage[pdftex]{graphicx}
\else
\fi
\usepackage{amsmath}
\usepackage{algorithmic}
\usepackage{array}

\usepackage{stfloats}
\usepackage{url}
\usepackage{hyperref}
\makeatletter
\def\UrlAlphabet{%
      \do\a\do\b\do\c\do\d\do\e\do\f\do\g\do\h\do\i\do\j%
      \do\k\do\l\do\m\do\n\do\o\do\p\do\q\do\r\do\s\do\t%
      \do\u\do\v\do\w\do\x\do\y\do\z\do\A\do\B\do\C\do\D%
      \do\E\do\F\do\G\do\H\do\I\do\J\do\K\do\L\do\M\do\N%
      \do\O\do\P\do\Q\do\R\do\S\do\T\do\U\do\V\do\W\do\X%
      \do\Y\do\Z}
\def\UrlDigits{\do\1\do\2\do\3\do\4\do\5\do\6\do\7\do\8\do\9\do\0}
\g@addto@macro{\UrlBreaks}{\UrlOrds}
\g@addto@macro{\UrlBreaks}{\UrlAlphabet}
\g@addto@macro{\UrlBreaks}{\UrlDigits}
\makeatother
\begin{document}
%
% paper title
% Titles are generally capitalized except for words such as a, an, and, as,
% at, but, by, for, in, nor, of, on, or, the, to and up, which are usually
% not capitalized unless they are the first or last word of the title.
% Linebreaks \\ can be used within to get better formatting as desired.
% Do not put math or special symbols in the title.
\title{Deep Multi-task Learning for Facial Expression Recognition and Synthesis Based on Selective Feature Sharing}

% author names and affiliations
% use a multiple column layout for up to three different
% affiliations
\author{\IEEEauthorblockN{Rui Zhao, Tianshan Liu, Jun Xiao, Daniel P.K. Lun, and Kin-Man Lam}
\IEEEauthorblockA{Department of Electronic and Information Engineering\\
The Hong Kong Polytechnic University, Hong Kong\\
Email: \{rick10.zhao, tianshan.liu, jun.xiao\}@connect.polyu.hk, \{enpklun, enkmlam\}@polyu.edu.hk}
}

% make the title area
\maketitle

% As a general rule, do not put math, special symbols or citations
% in the abstract
\begin{abstract}
Multi-task learning is an effective learning strategy for deep-learning-based facial expression recognition tasks. However, most existing methods take into limited consideration the feature selection, when transferring information between different tasks, which may lead to task interference when training the multi-task networks. To address this problem, we propose a novel selective feature-sharing method, and establish a multi-task network for facial expression recognition and facial expression synthesis. The proposed method can effectively transfer beneficial features between different tasks, while filtering out useless and harmful information. Moreover, we employ the facial expression synthesis task to enlarge and balance the training dataset to further enhance the generalization ability of the proposed method. Experimental results show that the proposed method achieves state-of-the-art performance on those commonly used facial expression recognition benchmarks, which makes it a potential solution to real-world facial expression recognition problems.
\end{abstract}

% no keywords

% For peer review papers, you can put extra information on the cover
% page as needed:
% \ifCLASSOPTIONpeerreview
% \begin{center} \bfseries EDICS Category: 3-BBND \end{center}
% \fi
%
% For peerreview papers, this IEEEtran command inserts a page break and
% creates the second title. It will be ignored for other modes.
\IEEEpeerreviewmaketitle

\section{Introduction}
\label{sec:intro}

Facial expression is one of the most effective and natural approaches for human beings to express their emotions. Identifying and synthesizing facial expressions enable current human-computer interaction systems to better understand and simulate human behaviors. In spite of their great effectiveness, the facial expression recognition algorithms, based on deep neural networks, are highly reliant on large amounts of training samples with clean labels. The annotation of a large-scale database is not only time consuming, but also impractical. To overcome this limitation, a number of studies \cite{Meng2017,Pons2018, Ranjan2017} have focused on multi-task learning for facial expression recognition (FER), which aims to train the models under the regularization from some auxiliary tasks. With the regularization effect, multi-task learning (MTL) becomes an effective learning strategy to tackle the overfitting problem, resulting from insufficient training samples. Moreover, learning multiple tasks simultaneously in a network can improve the performance by transferring beneficial knowledge to the main task from other relevant auxiliary tasks. Specifically, in FER, single-task networks are able to learn some very discriminative features with respect to facial expressions. However, the learnt features do not take the nuisance factors, such as subject identity, head pose, and illumination, into sufficient consideration, which results in poor generalization when applied to practical applications. Therefore, MTL contributes significantly to a more robust solution with better generalization for FER tasks.

There are two main underlying problems in current MTL-based FER algorithms, i.e. the design of the auxiliary tasks, and the building of the connections between different tasks. In terms of the auxiliary task design, numerous studies have been proposed for setting the tasks, such as expression-related hidden unit detection \cite{ReedLearningTD2014}, identity classification \cite{ZhangFacialER2017}, and landmark localization \cite{DevriesMulti-taskLF2014}, as the auxiliary tasks. In this paper, we propose to learn the facial expression synthesis (FES) as the auxiliary task for FER. FES aims to synthesize a facial expression image based on a guiding expression label. We employ a patch-based conditional generative adversarial network (cGAN) \cite{isola2017image} to learn the FES task. In addition to the high correlation with FER, FES can also generate extra training samples and balance the training dataset, which can greatly enhance the performance of a deep facial expression recognition framework.
 
On the other hand, establishing task interaction is another important factor when building multi-task networks, because the interaction will directly affect the information flow between different tasks. Conventional algorithms apply the hard parameter-sharing approach, which shares the feature maps at the bottom layers of a network and separates different branches for different tasks at the top layers, such as the two-head structure in object detection for classification and localization \cite{Ren-2015-FRT}. In spite of its simplicity, the hard parameter-sharing approach lacks the ability in differentiating helpful and harmful information between tasks. To address this issue, we propose a novel multi-task network for FER and FES, namely facial expression recognition and synthesis network (FERSNet), with a soft parameter-sharing mechanism, which contributes to effectively selecting useful features from different tasks and different layers. Therefore, the main contributions of this paper can be summarized as follows:

\begin{figure*}[t]
    \centering
    \includegraphics[width = 0.65\textwidth]{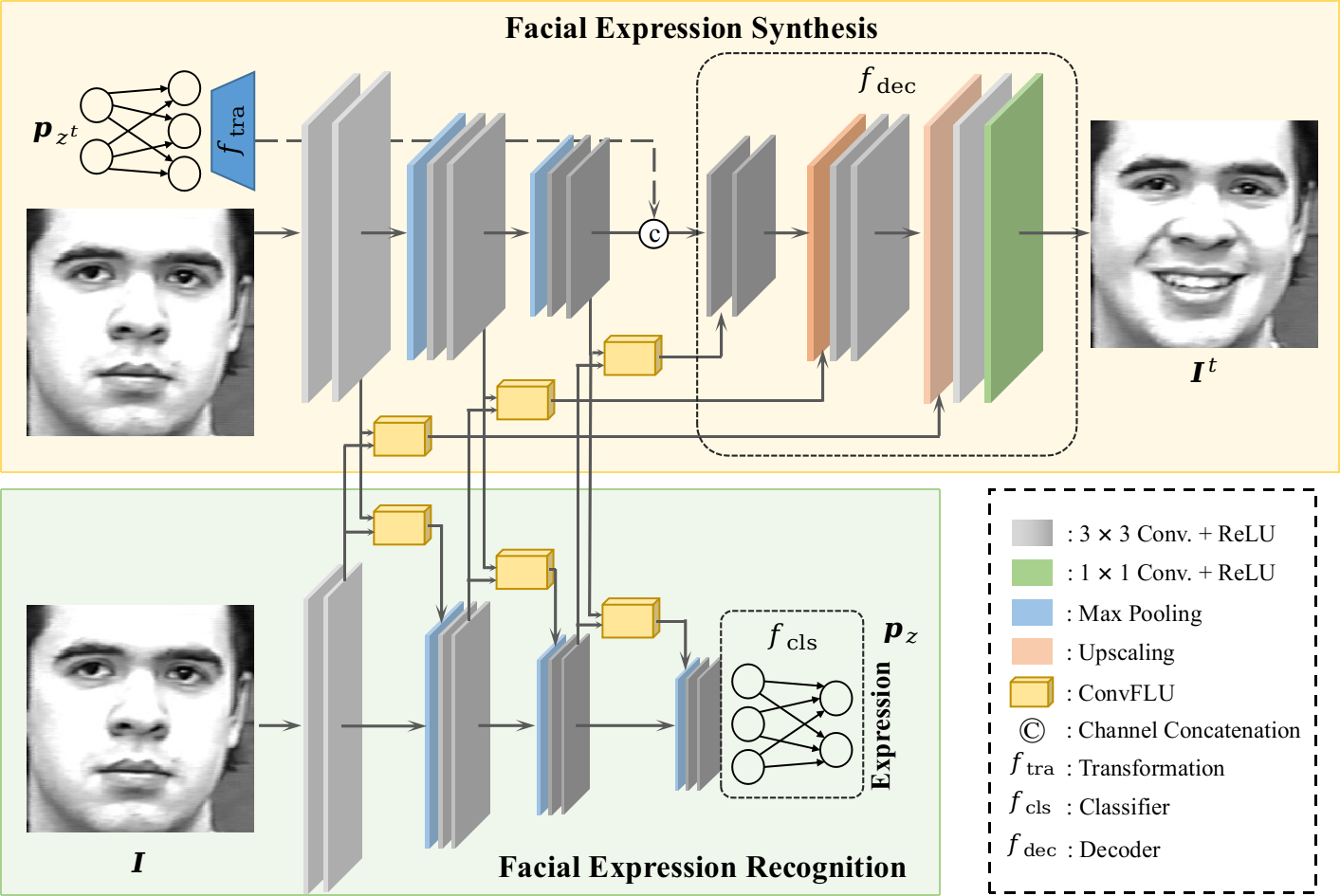}
    \caption{The proposed two-stream multi-task network for FER and FES with the convolutional feature leaky units (ConvFLU). $\bm{p}_{\hat{z}}$ is the output probability vector indicating the predicted expression label $\hat{z}$ of the input image, and $\bm{p}_{z^{t}}$ is a one-hot vector, which controls the expression label $z^{t}$ of the synthetic facial image. $E$ is the number of the expression classes.}
    \label{fig:1}
\end{figure*}

\begin{itemize}
    \item We propose a novel multi-task convolutional neural network, with the convolutional feature leaky unit, to selectively transfer the beneficial features between the facial expression recognition task and the facial expression synthesis task.
    \item We employ the facial expression synthesis branch to enlarge and balance the training dataset for further improving the generalization ability of the proposed algorithm.
    \item We conduct experiments to demonstrate that the proposed multi-task network achieves promising performance in recognizing and synthesizing facial expression images.
\end{itemize}

\section{Related Work}
\subsubsection{Multi-task learning for FER}
Multi-task learning for FER has been widely studied over the past few decades. Previous works on multi-task-based FER attempt to combine FER with other facial image analysis tasks, in order to obtain a more robust representation of facial expressions in the feature space. Meng et al. \cite{Meng2017} proposed a two-stream network to extract identity-invariant expression features for the emotion classification. Pons and Masip \cite{Pons2018} suggested that jointly learning a model for FER and facial action units detection can significantly improve the FER performance. Moreover, Ranjan et al. \cite{Ranjan2017} proposed a multi-branch network to solve diverse facial image analysis tasks simultaneously. Zhang et al. \cite{ZhangFacialER2017} proposed a multi-signal CNN under the supervision of the FER and face verification tasks, which forces the model to learn more discriminative features with respect to facial expressions. Ming et al. \cite{DMTL2019} proposed a multi-task network with the dynamic weights for the FER and face recognition tasks to enhance the model performance. However, the above-mentioned studies have not considered feature selection when sharing information between different tasks, which may greatly degrade their performance, because useless, or even harmful information, is transferred.

\subsubsection{Facial expression synthesis}
Facial expression synthesis is another widely studied topic in the field of facial image analysis. With the development of generative adversarial network (GAN) \cite{NIPS2014_5423}, the facial expression edit/synthesis has achieved appealing performance. Zhang and Song \cite{zhang2017age} proposed a conditional adversarial autoencoder (CAAE) for synthesizing the facial images with different expressions and ages. ExprGAN \cite{ding2017exprgan} adopted the conditional GAN strategy to produce facial images with different expression intensities. Moreover, Choi et al. \cite{8579014} proposed StarGAN to achieve multi-domain image-to-image translation for facial image synthesis. In addition, the geometric-guided methods \cite{Geo1,Geo2,Geo3} employed the shape-aware supervision from the facial landmarks for expression editing, which achieved state-of-the-art performance in facial expression transference. However, limited existing work employs the soft parameter-sharing strategy to enhance the quality of the synthetic facial images with the FER regularization.

\subsubsection{Feature selection mechanism}
The selection mechanism plays an important role in multi-task learning, which has been widely applied to natural language processing (NLP). Ruder et al. \cite{Ruder2017SluiceNL} presented Sluice Networks, in which a subspace combination approach was proposed to determine the information flow between different tasks. Moreover, Xiao et al. \cite{xiao-etal-2018-learning} took advantage of gated recurrent unit (GRU) and proposed a leaky unit with the property of remembering and forgetting information, which achieved state-of-the-art performance in text classification. However, different from the NLP tasks, the facial image analysis tasks usually fuse the local features to form a global representation of the image. Therefore, in this paper, we propose a convolutional feature leaky unit (ConvFLU) to perform feature selection between different tasks and layers.

\section{The Proposed Method}
The proposed FERSNet solves the FER and FES tasks in parallel with the reliable knowledge transference. Fig. \ref{fig:1} illustrates the pipeline of FERSNet. The branch at the top aims to recognize the expression of the input facial image, while the branch at the bottom aims to generate a new facial expression image with the same identity as the input, based on the target-expression label. The two branches are connected by a set of soft parameter-sharing blocks, i.e. ConvFLUs. In this section, we first introduce the proposed multi-task framework for the FER and FES tasks, and then we present the details of ConvFLU for selective feature sharing. Finally, we show the learning strategy for the proposed multi-task network to solve the FER and FES problems simultaneously.

\subsection{Framework}
As presented in Fig. \ref{fig:1}, the proposed two-stream network consists of two branches for FER and FES, respectively. The first four convolutional blocks in the FER and FES branches are connected with ConvFLUs, which aim to extract discriminative features for the FER and FES tasks, respectively. Based on the extracted FER features, the classifier $f_{\text{cls}}$ predicts the emotion label $\hat{z}$ of the input facial image. In the FES branch, another input variable $z^{t}$, which controls the expression of the synthetic image, is fed to the transformer $f_{\text{tra}}$. $f_{\text{tra}}$ encodes the information of the target-expression label to produce a feature map with the same size as the extracted FES features. The two feature maps are fused by channel concatenation. Finally, the decoder $f_{\text{dec}}$ reconstructs the image based on the fused features to mislead a patch-based discriminator \cite{isola2017image}.

In summary, the proposed FERSNet takes a facial image and a target-expression label as inputs to predict the expression label of the input image and synthesize another facial image with the expected (target) expression.

\subsection{Convolutional Feature Leaky Unit}
\begin{figure}[t]
    \centering
    \includegraphics[width = 0.75\linewidth]{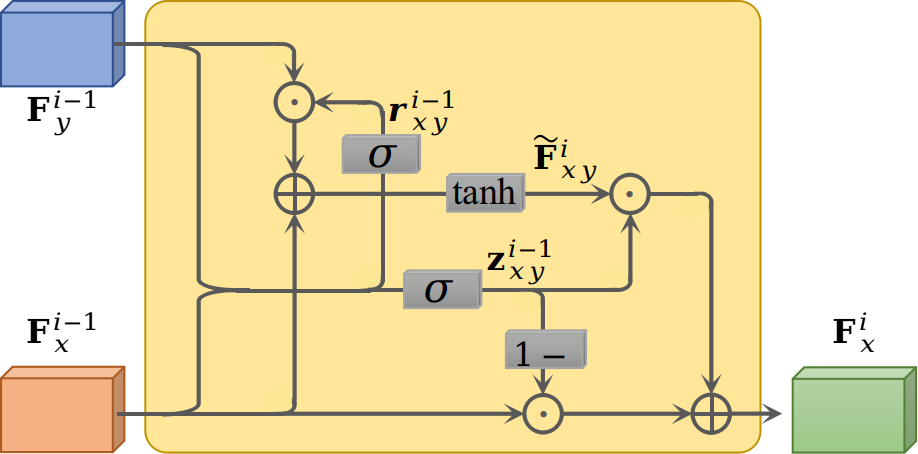}
    \caption{The proposed ConvFLU in FERSNet. $\odot$ and $\oplus$ represent the element-wise multiplication and addition, respectively. $\sigma$ and ``tanh'' refer to the sigmoid function and the hyperbolic tangent function, respectively.}
    \label{fig:2}
\end{figure}
Inspired by GRU in \cite{cho-2014-learning}, we propose a convolutional feature leaky unit, which inherits the effective property of remembering and forgetting features. We revise the original structure of GRU and employ it to filter out useless and harmful features, when transferring information between the different tasks. The structure of the proposed ConvFLU is illustrated in Fig. \ref{fig:2}. We consider transferring the features from task $y$ to task $x$ in the $(i-1)$-st layer. It is worth noting that there are two ConvFLUs in one transference block, as shown in Fig. \ref{fig:1}. Therefore, the information flow is bidirectional between FER and FES.

\begin{figure}[t]
    \centering
    \includegraphics[width = 0.99\linewidth]{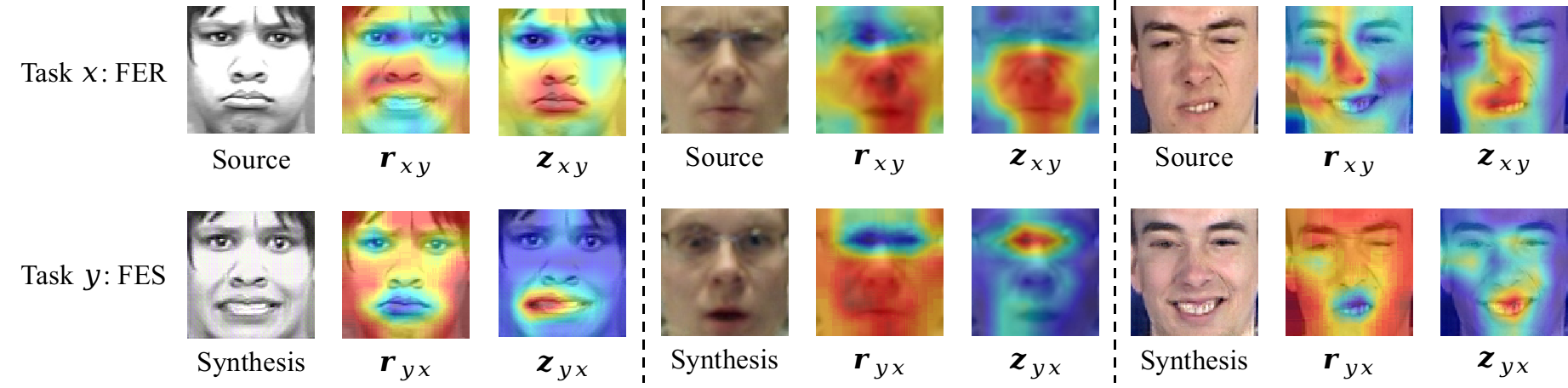}
    \caption{The visualization of the gate values in ConvFLU. The blue and red values represent the low and high response values, respectively. Task $x$ and $y$ are the FER and the FES task in FERSNet, respectively.}
    \label{fig:vg}
\end{figure}

The leaky gate $\mathbf{r}_{xy}^{i-1}$ for transferring information from task $y$ to task $x$ is defined as follows:
\begin{equation}
    \mathbf{r}_{xy}^{i-1} = \sigma(\mathbf{W}_{\mathbf{r}}^{i-1} \ast [\mathbf{F}_{x}^{i-1}, \mathbf{F}_{y}^{i-1}]),
\end{equation}
where $\mathbf{W}_{\mathbf{r}}^{i-1}$ represents the trainable convolutional kernels for $\mathbf{r}_{xy}^{i-1}$. ``[ ]'' and ``$\ast$'' denote the concatenation and the convolution operation, respectively. $\mathbf{F}_{x}^{i-1}$ and $\mathbf{F}_{y}^{i-1}$ are the input feature maps from task $x$ and task $y$ in the $(i-1)$-st layer, respectively. Then, a new feature map is generated based on $\mathbf{r}_{xy}^{i-1}$ as follows:
\begin{equation}
    \widetilde{\mathbf{F}}_{xy}^i = \text{tanh}(\mathbf{W}^{i-1} \ast (\mathbf{r}_{xy}^{i-1} \odot \mathbf{F}_{y}^{i-1}) + \mathbf{U}^{i-1}\ast \mathbf{F}_{x}^{i-1}),
\end{equation}
where $\mathbf{U}^{i-1}$ and $\mathbf{W}^{i-1}$ are the trainable convolutional kernels for combining $\mathbf{F}_{x}^{i-1}$ and $\mathbf{F}_{y}^{i-1}$. $\odot$ denotes the element-wise multiplication. It is clear that $\mathbf{r}_{xy}^{i-1}$ controls the information leakage from task $y$ to task $x$. Moreover, we further consider a memory gate $\mathbf{z}_{xy}^{i-1}$, which determines the information that should be remembered from the previous feature map. $\mathbf{z}_{xy}^{i-1}$ is defined as follows:
\begin{equation}
    \mathbf{z}_{xy}^{i-1} = \sigma(\mathbf{W}_{\mathbf{z}}^{i-1} \ast [\mathbf{F}_{x}^{i-1},\mathbf{F}_{y}^{i-1}]),
\end{equation}
where $\mathbf{W}_{\mathbf{z}}^{i-1}$ is the trainable convolutional kernels for generating $\mathbf{z}_{xy}^{i-1}$. Thus, the final feature map $\mathbf{F}_{x}^{i}$, containing both the information from $\mathbf{F}_{x}^{i-1}$ and $\mathbf{F}_{y}^{i-1}$, is computed as follows:
\begin{equation}
    \mathbf{F}_{x}^{i} = (1 - \mathbf{z}_{xy}^{i-1}) \odot \mathbf{F}_{x}^{i-1} + \mathbf{z}_{xy}^{i-1} \odot \widetilde{\mathbf{F}}_{xy}^{i}.
\end{equation}
With the leaky gate and the memory gate, ConvFLU is able to select the beneficial features for FER and FES from each other. If the values in the leaky gates $\mathbf{r}_{xy}$ are close to $1$, the branch tends to utilize more information from the other task. Similarly, if the values in the memory gates $\mathbf{z}_{xy}$ are close to $0$, the branch tends to preserve its information for the corresponding task. We further visualize the leakage gate and the memory gate in Fig. \ref{fig:vg} to better illustrate the selective feature-sharing strategy. 
It can be seen from the figure that the FER task (task $x$) mainly focuses on the mouth and the eye regions to predict the expression label, while the FES task (task $y$) tends to preserve the identity information, when synthesizing a new expression. Therefore, the proposed ConvFLU can be regarded as an attention mechanism, based on task correlation, for performing selective feature sharing.

\subsection{Learning for FER and FES}
In this paper, we solve FER and FES simultaneously with the proposed FERSNet. For the FER task, the classifier in the top branch consists of fully connected layers, under a standard cross entropy loss for training, which is defined as follows:
\begin{equation}
    \mathcal{L}_{\text{cls}} = \frac{1}{n}\sum_{i=1}^{n} -\text{log}\left( \frac{\text{exp}(\bm{p}_{i}[z_i])}{\sum_{j}\text{exp}(\bm{p}_i[j])}\right),
\end{equation}
where $\bm{p}_i$ denotes the output probability vector of the $i$-th training sample, $z_i$ represents its corresponding ground-truth label, and $n$ denotes the total number of training samples.

In terms of the FES task in the bottom branch, we consider the loss from a patch-based discriminator \cite{isola2017image}, denoted as $\mathcal{L}_{\text{GAN}}$, and also adopt the reconstruction loss, $\mathcal{L}_{\text{rec}}$, to synthesize visually pleasant facial images. Specifically, the GAN loss is defined as follows:
\begin{equation}
    \begin{aligned}
        \mathcal{L}_{\text{GAN}} = & \mathbb{E}_{z^{t}, \bm{I}^{t}}[\text{log}D(z^{t}, \bm{I}^{t})]+ \\
        & \mathbb{E}_{z^{t}, \bm{I}}[\text{log}(1-D(z^{t}, G(z^{t}, \bm{I})))],
    \end{aligned}
\end{equation}
where $\bm{I}^t$ and $\bm{I}$ denote the real target and the original facial image, respectively. $z^t$ represents the target-expression label. $G$ and $D$ refer to the generator and the discriminator in the network, respectively. Specifically, the generator consists of all the modules in FERSNet for producing the synthetic face. As for the reconstruction loss, we define it as the mean squared error between the target and the synthetic facial image as follows:
\begin{equation}
    \mathcal{L}_{\text{rec}} = \frac{1}{n}\sum_{i=1}^{n}||G(z^{t}, \bm{I}_{i}) - \bm{I}^{t}_{i}||^{2}_{2}.
\end{equation}
In addition, we follow the learning strategy in \cite{Geo1,Geo2,Geo3}, and employ the cycle consistency loss and the identity preserving loss to further improve the synthesis accuracy. The cycle consistency loss $\mathcal{L}_{\text{cyc}}$ guarantees the consistency between the source image and the cycle-reconstructed image, which is defined as:
\begin{equation}
    \label{eq:cyc}
    \mathcal{L}_{\text{cyc}} = \frac{1}{n} \sum_{i=1}^{n} ||G(z_i, G(z^{t}, \bm{I}_{i})) - \bm{I}_{i}||^2_2.
\end{equation}
In other words, if we transform the synthetic face back to the original expression $z_i$ via FERSNet, the same facial image as $\bm{I}_i$ should be obtained. Moreover, the identity preserving loss $\mathcal{L}_{\text{idt}}$ is defined as the distance between the features extracted from the original face and the synthetic face with the pre-trained model-B of the Light CNN \cite{Wu2015ALC}, which is formulated as follows:
\begin{equation}
    \label{eq:idt}
    \mathcal{L}_{\text{idt}} = \frac{1}{n}\sum_{i=1}^{n}||f_{\text{LiCNN}}(G(z^{t}, \bm{I}_{i})) - f_{\text{LiCNN}}(\bm{I}^{t}_{i})||^{2}_{2},
\end{equation}
where $f_{\text{LiCNN}}$ denotes the pre-trained Light CNN for feature extraction. As the Light CNN aims to recognize the identity information based on the facial images, it can extract the most prominent features for identity discrimination. 

Therefore, the overall loss function for training FERSNet is computed as follows:
\begin{equation}
\label{eq:all}
    \mathcal{L} = \mathcal{L}_{\text{cls}} + \lambda_{1}\mathcal{L}_{\text{GAN}} + \lambda_{2}\mathcal{L}_{\text{rec}} + \lambda_3 \mathcal{L}_{\text{cyc}} + \lambda_4 \mathcal{L}_{\text{idt}},
\end{equation}
where $\lambda_{1}$, $\lambda_{2}$, $\lambda_{3}$, and $\lambda_{4}$ are the hyperparameters, controlling the trade-off between the different loss terms.
\section{Experiments}
In this section, we present the implementation details and the experiment settings for evaluating FERSNet. We also compare FERSNet with other state-of-the-art methods for facial expression recognition and synthesis.
\subsection{Implementation Details}
\label{Sec:ID}
As shown in Fig. \ref{fig:1}, the FER branch employs four convolutional blocks. Each convolutional block consists of two convolutional layers, two batch normalization \cite{BN} layers, two ReLU \cite{ReLU} activation layers, and one max pooling layer. For the FES branch, we employ eights convolutional blocks. The first four blocks are established with the same structure as that of the FER branch. The next four blocks form the decoder $f_{\text{dec}}$ in Fig. \ref{fig:1}, in which we replace the max pooling layer with the deconvolutional layer to upscale the feature maps for reconstructing the target image. The classifier $f_{\text{cls}}$ consists of two fully connected layers, which produces a $E$-dimensional logit, indicating the predicted emotion probability. The transformer $f_{\text{tra}}$ makes an inverse mapping to generate a feature map from the target $E$-dimensional logit, and thus it consists of fully connected layers and deconvolutional layers. In the network, all the convolutional kernels are of size $3\times3$, with padding $1$ and stride $1$, except for those in ConvFLUs, where we utilize convolutional filters of size $1\times 1$, with padding $0$ and stride $1$. The max pooling layers and the deconvolutional layers consist of $2 \times 2$ kernels with stride $2$ for rescaling the feature maps.

In the training phase, we randomly select a target label $z^{t}_i$ with the corresponding facial image $\bm{I}^{t}_{i}$ for each training sample. We apply face alignment to all the facial images, based on the method \cite{bulat2017far}. Each aligned face is resized to $110 \times 110$. We then randomly crop a $96 \times 96$ region from the aligned facial images, and apply random mirroring and random rotation at $\{ -15^\circ, -10^\circ, -5^\circ, 0^\circ, 5^\circ, 10^\circ, 15^\circ\}$ to the cropped images to obtain the final training samples. For each testing image, we apply the same face alignment method \cite{bulat2017far}, and resize the aligned face to $110 \times 110$. We adopt the center-crop approach to generate the final testing sample with size $96 \times 96$ from each testing image. It is worth noting that the above-mentioned pre-processing and augmentation approaches are commonly used in those methods \cite{8038215,Jung_2015_ICCV,Yang2018,LBVCNN} compared in our experiments.

We implement FERSNet with PyTorch \cite{paszke2017automatic}. During training, we set the batch size to $64$, and the learning rate is set to decrease from $10^{-3}$ to $10^{-5}$ within $500$ epochs based on the cosine annealing strategy \cite{loshchilov2016sgdr}. We adopt the Adam \cite{Adam} optimizer to minimize the objective function, defined in Eq. (\ref{eq:all}), with $\lambda_1$, $\lambda_2 $, and $\lambda_3$ empirically set to $0.3$, $1$, and $0.5$, respectively. $\lambda_4$ is set to 0.1 at the beginning, and is gradually increased to $0.5$ during the training process, which also follows the design in \cite{Geo1}. We train the network on a Nvidia GEFORCE GTX 1080 Ti GPU, and it takes about 5 hours to train up one FERSNet model.

\subsection{Evaluation on FER}
\begin{table}[t]
    \centering
    \caption{The number of \textbf{video sequences} in CK+, Oulu-CASIA, and MMI, based on different emotion labels.}
    \begin{tabular}{|c|c c c c c c c|c|}
    \hline
        Database & An & Co & Di & Fe & Ha & Sa & Su & Total\\
        \hline
        CK+ & 45 & 18 & 59 & 25 & 69 & 28 & 83 & 327 \\
        Oulu-CASIA & 80 & - & 80 & 80 & 80 & 80 & 80 & 480 \\
        MMI & 33 & - & 32 & 28 & 42 & 32 & 41 & 208 \\
        \hline
    \end{tabular}
    \label{tab:0}
\end{table}
As the proposed FERSNet aims to recognize and synthesize facial expression images, the training dataset is required to contain both the expression and the identity information. Therefore, we employ three commonly used facial expression benchmarks, which are the the Extended Cohn-Kanade dataset (CK+) \cite{KanadeCK+}, the Oulu-CASIA NIR\&VIS facial expression database (Oulu-CASIA) \cite{ZhaoOulu}, and the MMI facial expression database (MMI) \cite{1521424}, to evaluate the emotion recognition performance of FERSNet. In addition, we mainly consider six standard facial expressions, i.e. anger (An), disgust (Di), fear (Fe), happiness (Ha), sadness (Sa) and surprise (Su) in our experiments, because these six expressions are universal among humans, irrespective of their age, gender and race \cite{FACS}. Moreover, one additional emotion class, i.e. contempt (Co), is included, when evaluating FERSNet on CK+. The number of video sequences in each database is summarized in terms of the expression labels in Table \ref{tab:0}.

The \textbf{Extended Cohn-Kanade} (CK+) dataset \cite{KanadeCK+} consists of 593 video sequences collected from 123 subjects. We use the video sequences with the provided seven expression labels, and select the last three peak frames as the emotional faces, which results in 981 images in total. We further split the images into 10 folds based on the identity, and perform the 10-fold identity-independent cross-validation to train FERSNet using 90\% of the samples and test its performance using the remaining 10\% of the samples. The final recognition results are obtained by averaging the accuracy over the 10 runs.

\begin{table}[t]
    \centering
    \caption{The recognition accuracy of the seven expressions on CK+ based on different methods. The best results are highlighted in \textbf{bold}.}
    \begin{tabular}{|c|c|c|c|}
    \hline
        Methods & Pre-train & Setting & Accuracy (\%) \\
        \hline
        LBP-TOP \cite{zhao2007dynamic} & \xmark & Image sequence & 88.99\\
        HOG 3D \cite{klaser2008} & \xmark & Image sequence & 91.44\\
        3DCNN \cite{3DCNN} & \xmark  & Image sequence & 85.9 \\
        IACNN \cite{8038215}  & \cmark & Single image & 95.37 \\
        DTAGN \cite{Jung_2015_ICCV} & \cmark & Image sequence & 97.25 \\
        IPA2LT \cite{Zeng_2018_ECCV} & \xmark & Single image & 91.67 \\
        DeRL \cite{Yang2018} & \cmark & Single image & 97.30 \\
        LBVCNN \cite{LBVCNN} & \cmark & Image sequence & 97.38 \\
        DMT-CNN \cite{DMTL2019} & \cmark & Single image & 97.55 \\
        \hline
        FERSNet & \xmark & Single image & 97.35 \\
        FERSNet (BU-4DFE) & \cmark & Single image & \textbf{97.85} \\
        \hline
    \end{tabular}
    \label{tab:1}
\end{table}
The comparison results on CK+ are listed in Table \ref{tab:1}. As DeRL \cite{Yang2018} was pre-trained on the BU-4DFE dataset \cite{6553788}, we established two versions of FERSNet, which are trained from scratch and pre-trained on BU-4DFE, to make the comparison with DeRL fair and reasonable. The BU-4DFE dataset \cite{6553788} contains $60,600$ images from 101 subjects with the six standard emotion labels. We follow the settings in Sec. \ref{Sec:ID} to pre-train FERSNet on BU-4DFE. In addition, the original DMT-CNN is designed for recognizing the eight facial expressions, i.e the original seven expressions and the neutral faces, in CK+. To make a fair comparison, we established and re-trained the DMT-CNN model, based on the settings in \cite{DMTL2019}, to recognize the the original seven expressions in CK+. It can be observed from the table that the proposed FERSNet outperforms all the other methods on CK+. Compared to DeRL, which is also a ``GAN + Classifier'' method for FER, FERSNet obtains an accuracy improvement of about 0.55\%.

The \textbf{Oulu-CASIA} \cite{ZhaoOulu} database consists of video sequences under three different illumination conditions. In our experiments, we only use 480 video sequences, taken from 80 subjects under the strong illumination condition. There are six emotion labels in Oulu-CASIA, i.e. anger (An), disgust (Di), fear (Fe), happiness (Ha), sadness (Sa) and surprise (Su). For each video sequence, we select the last three peak frames with the provided emotion label to form the dataset, which results in $1,440$ images in total. Similar to CK+, the 10-fold identity-independent cross-validation is performed to evaluate FERSNet on Oulu-CASIA.
\begin{table}[ht]
    \centering
    \caption{The recognition accuracy of the six expressions on Oulu-CASIA based on different methods. The best results are highlighted in \textbf{bold}.}
    \begin{tabular}{|c|c|c|c|}
    \hline
        Methods & Pre-train & Setting & Accuracy (\%) \\
        \hline
        LBP-TOP \cite{zhao2007dynamic} & \xmark & Image sequence & 68.13\\
        HOG 3D \cite{klaser2008} & \xmark & Image sequence & 70.63\\
        STM-Explet \cite{6909622} & \xmark & Image sequence & 74.59\\
        DTAGN \cite{Jung_2015_ICCV} & \cmark & Image sequence & 81.46 \\
        IPA2LT \cite{Zeng_2018_ECCV} & \xmark & Single image & 61.02 \\
        DeRL \cite{Yang2018} & \cmark & Single image & 88.0 \\
        LBVCNN \cite{LBVCNN} & \cmark & Image sequence & 82.41 \\
        DMT-CNN \cite{DMTL2019} & \cmark & Single image & 87.5 \\
        ExprGAN \cite{ding2017exprgan} & \cmark & Single image & 84.72\\
        \hline
        FERSNet & \xmark & Single image & 83.47\\
        FERSNet (BU-4DFE) & \cmark & Single image & \textbf{89.23} \\
        \hline
    \end{tabular}
    \label{tab:2}
\end{table}

The comparison results on Oulu-CASIA are summarized in Table \ref{tab:2}. FERSNet pre-trained on BU-4DFE outperforms all the competitors, and it surpasses DeRL by about 1.2\%. In addition, we observe that with the pre-training on BU-4DFE, FERSNet acquires a larger accuracy improvement of about 6\% on Oulu-CASIA, and the FERSNet model trained from scratch achieves an accuracy of 83.47\%.

The \textbf{MMI} database \cite{1521424} consists of 236 video sequences, recorded from 31 subjects. Each video sequence is labelled as one of the six standard emotions. We select 208 video sequences with the frontal-view faces. Because the label is provided for each sequence and the peak expression face mainly appears in the middle, we further select the three frames in the middle of each sequence, which results in 624 images in total. We also follow the 10-fold identity-independent cross-validation strategy to evaluate the performance of FERSNet on MMI. The final accuracy is averaged over the 10 runs on MMI. We present the comparison results in Table \ref{tab:MMI}.

\begin{table}[ht]
    \centering
    \caption{The recognition accuracy of the six expressions on MMI based on different methods. The best results are highlighted in \textbf{bold}.}
    \resizebox{\linewidth}{!}{
    \begin{tabular}{|c|c|c|c|}
    \hline
        Methods & Pre-train & Setting & Accuracy (\%) \\
        \hline
        LBP-TOP \cite{zhao2007dynamic} & \xmark & Image sequence & 59.51\\
        HOG 3D \cite{klaser2008} & \xmark & Image sequence & 60.89 \\
        STM-Explet \cite{6909622} & \xmark & Image sequence & 75.12 \\
        DTAGN \cite{Jung_2015_ICCV} & \cmark & Image sequence & 70.24 \\
        IACNN \cite{8038215}  & \cmark & Single image & 71.55 \\
        DeRL \cite{Yang2018} & \cmark & Single image & 73.23 \\
        LBVCNN \cite{LBVCNN} & \cmark & Image sequence & \textbf{76.28} \\
        \hline
        FERSNet & \xmark & Single image & 71.31 \\
        FERSNet (BU-4DFE) & \cmark & Single image & 75.32 \\
        \hline
    \end{tabular}}
    \label{tab:MMI}
\end{table}
It can be seen from the table that the sequence-based method, i.e. LBVCNN \cite{LBVCNN}, achieves higher accuracy than the proposed FERSNet. Those sequence-based methods employ the temporal information, while FERSNet only considers the spatial information from a static image. We achieve about 75.3\% on MMI, which is very close to the sequence-based methods and still surpasses DeLR by about 2\%.

In the above experiments, FERSNet consistently outperforms DeLR on CK+, Oulu-CASIA, and MMI, which shows the effectiveness of jointly learning for FER and FES. Compared to the de-expression strategy in DeRL \cite{Yang2018}, the proposed facial expression editing (synthesizing) strategy is a more general case for transferring expression information, and thus FERSNet produces better results.

\subsection{Evaluation on FES}
\begin{figure}[ht]
    \centering
    \subfigure[CK+]{
    \centering
    \includegraphics[width=1\linewidth]{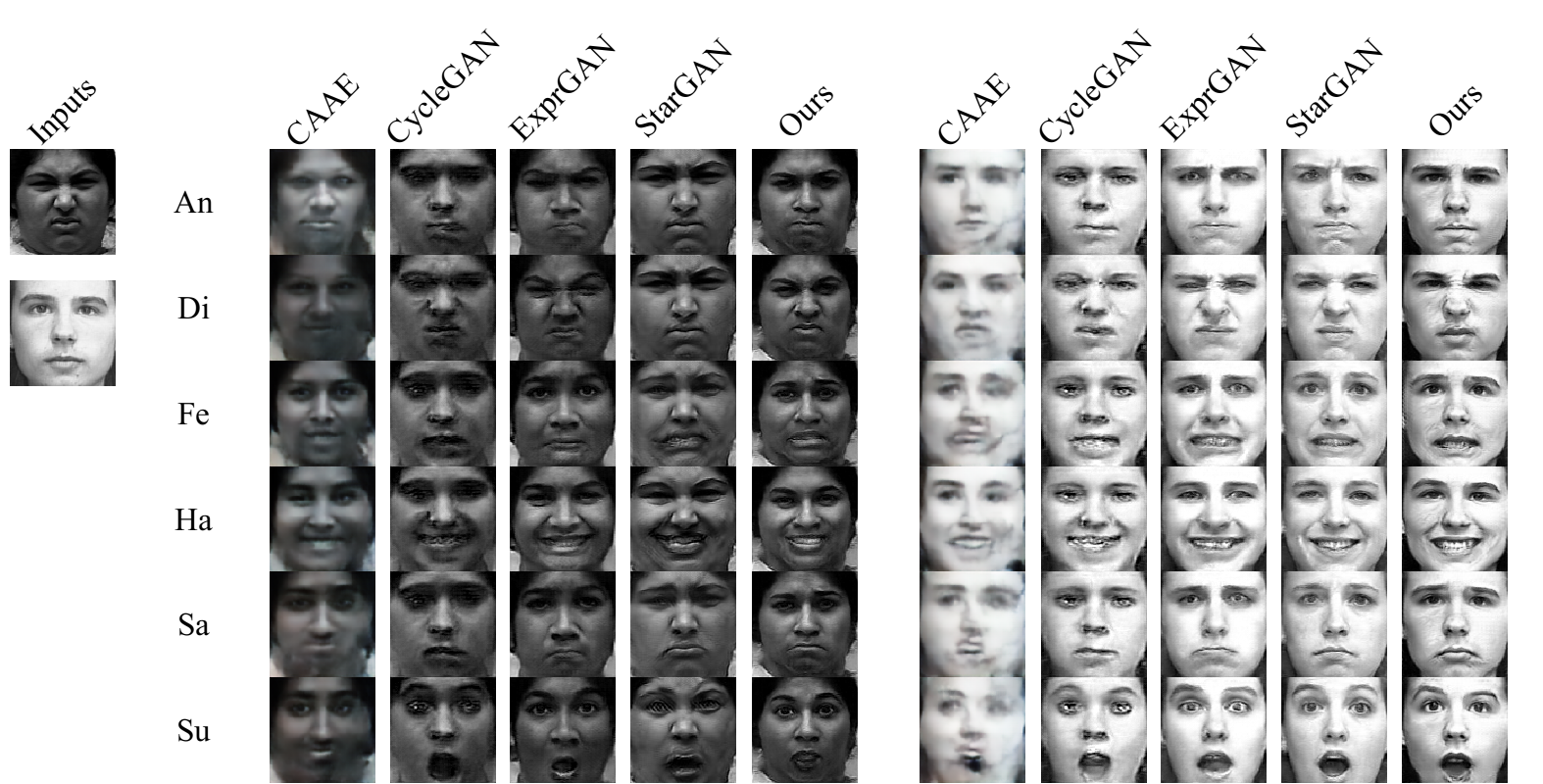}
    }
    \subfigure[Oulu-CASIA]{
    \centering
    \includegraphics[width=1\linewidth]{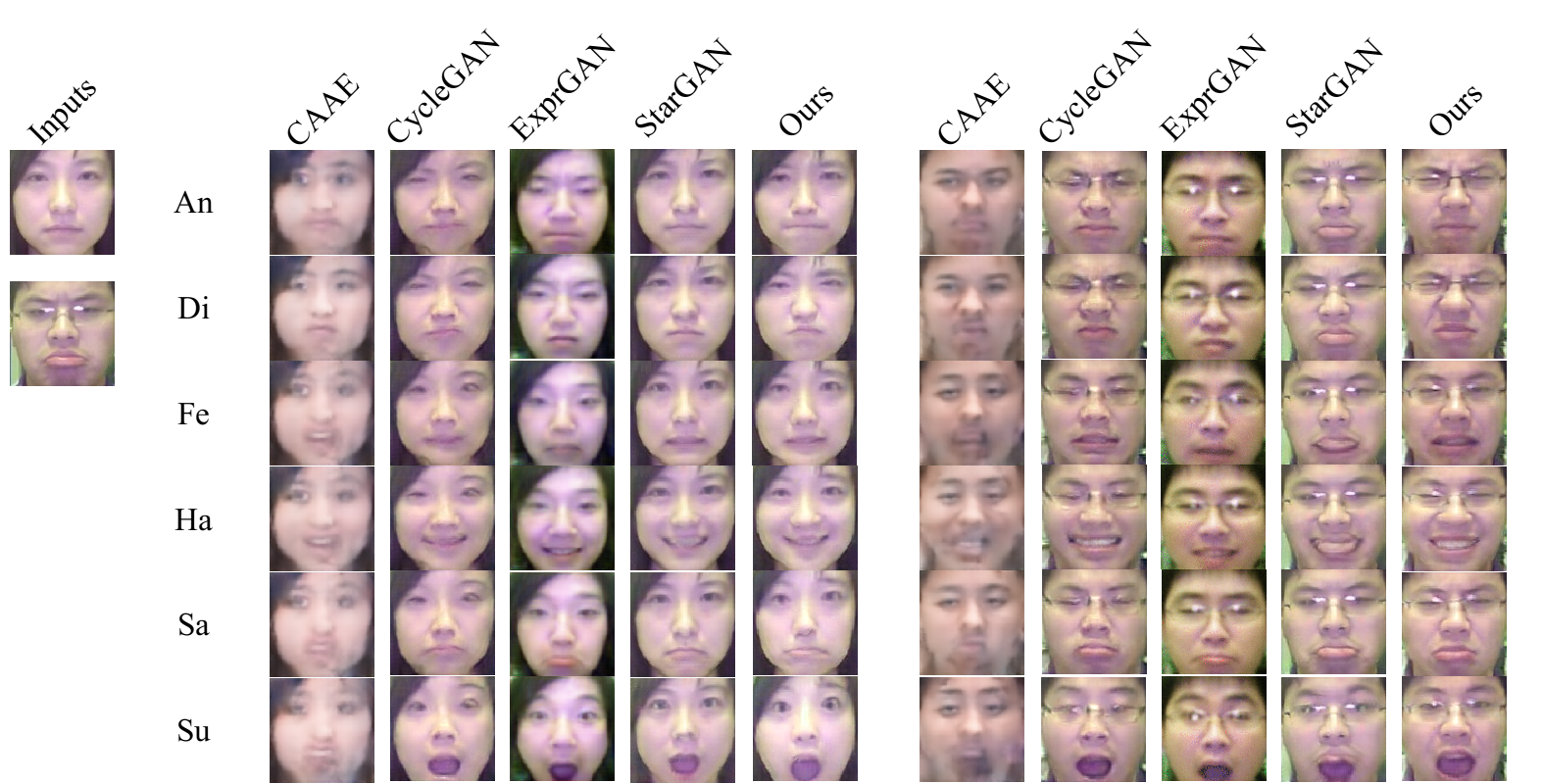}
    }
    \subfigure[MMI]{
    \centering
    \includegraphics[width=1\linewidth]{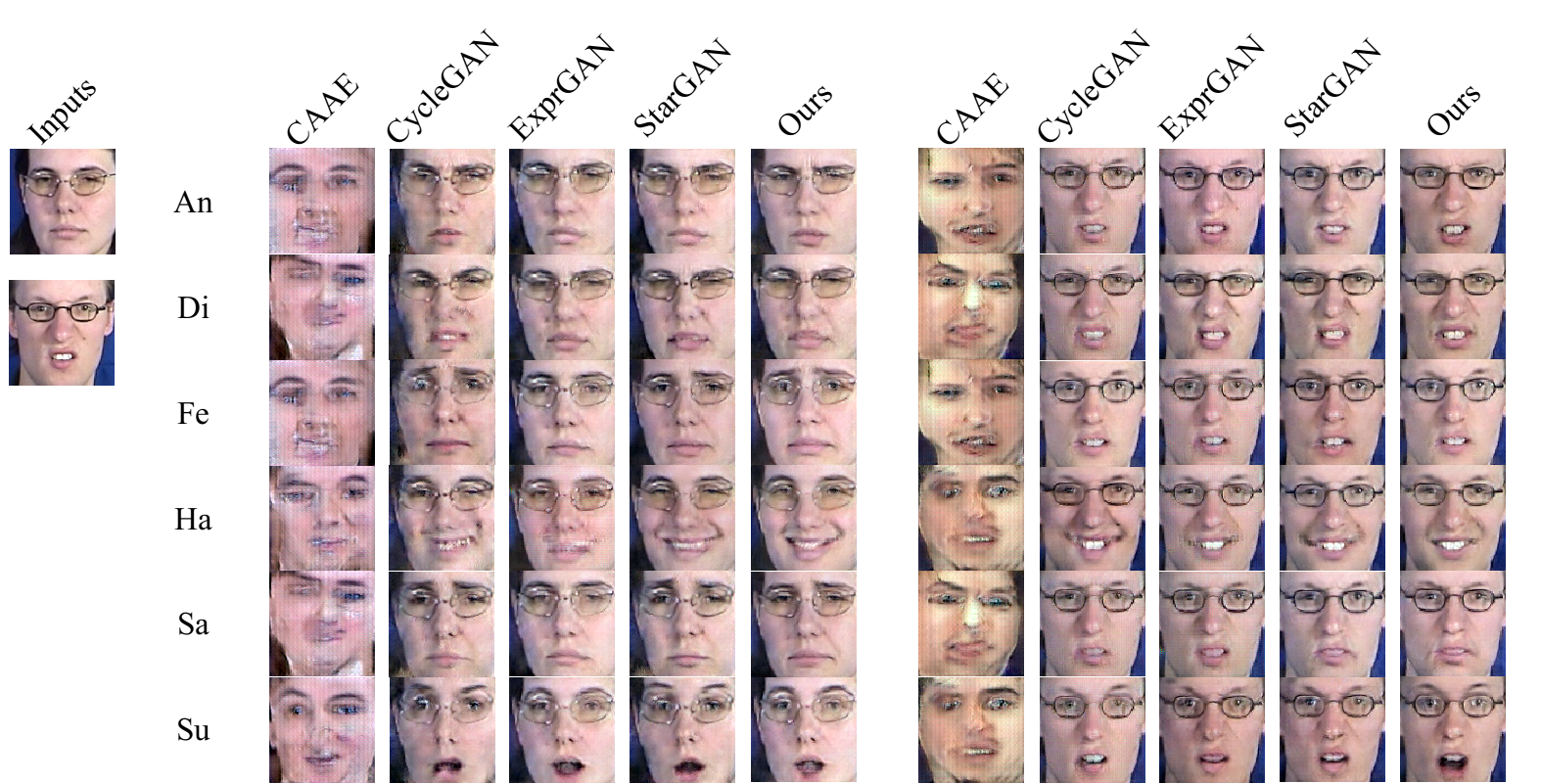}
    }
    \caption{The visual results of the facial expression synthesis task on CK+, Oulu-CASIA, and MMI, based on the different generative methods.}
    \label{fig:3}
\end{figure}
We evaluate the performance of FERSNet on synthesizing facial images with the expected expression. In the FES experiments, we mainly consider the six basic emotions, and we present the visual results of FERSNet with the other generative models, including StarGAN \cite{8579014}, ExprGAN \cite{ding2017exprgan}, CycleGAN \cite{CycleGAN2017} and CAAE \cite{zhang2017age} on CK+ \cite{KanadeCK+} and Oulu-CASIA \cite{ZhaoOulu}. Among them, ExprGAN and StarGAN are the conditional generative frameworks, which are specially designed for facial expression synthesis. It is worth noting that the geometric-guided methods \cite{Geo1,Geo2,Geo3} are not included in the comparison, because they obviously use much stronger supervision than the proposed algorithm. The qualitative results are presented in Fig. \ref{fig:3}, which clearly shows that CAAE and CycleGAN fail to produce satisfactory facial images, as they create the images with visible distortions. ExprGAN and StarGAN generally produce comparable results with our proposed method. Nevertheless, the proposed FERSNet can better maintain the color consistency on Oulu-CASIA. In addition, it can be observed from Fig. \ref{fig:3} that FERSNet exhibits a better ability to generate the eye and the mouth regions, and preserve the identity information. This is because the leaky gate and the memory gate in ConvFLU can more effectively transfer and memorize the related features for the FES task. In summary, FERSNet synthesizes more visually pleasant images with less artefacts and blurs.

To further validate FERSNet on the FES task, we follow the evaluation approach in StarGAN \cite{8579014} and ExprGAN \cite{ding2017exprgan}, and establish the quantitative comparison. Specifically, we train the above-mentioned FES methods on the training set, and then perform the expression synthesis on the unseen testing set. The synthetic facial images are fed to an expression recognition network. A higher recognition accuracy on the synthetic facial images indicates more realistic expression synthesis, because the generated images lie in the same manifold of natural expressions. It is worth noting that the expression recognition network is independently trained on the original training set, containing real facial images only. Then, we employ this network to recognize the synthetic images from the different generative models. The results are summarized in Table \ref{tab:FES}. We also adopt the 10-fold identity-independent cross-validation strategy in this experiment, and the final recognition results are obtained by averaging the accuracy over the 10 runs.
\begin{table}[ht]
    \centering
    \caption{The recognition accuracy (\%) on the synthetic facial images produced by the different methods. The best results are highlighted in \textbf{bold}.}
    \resizebox{1\linewidth}{!}{
    \begin{tabular}{|c|c c c c |c|}
        \hline
        Method & CAAE & CycleGAN & ExprGAN & StarGAN & Ours \\
        \hline
        CK+ & 79.41 & 88.89 & 95.41 & 96.43 & \textbf{97.04} \\
        Oulu-CASIA & 46.18 & 74.44 & 80.07 & 79.51 & \textbf{81.52} \\
        \hline
    \end{tabular}
    }
    \label{tab:FES}
\end{table}
As shown in the table, the proposed FERSNet achieves the highest accuracy, which demonstrates its superiority in synthesizing more realistic facial expression images.

\subsection{Ablation Study}
In order to make a comprehensive analysis on the proposed method, we present the ablation study to investigate the effectiveness of the different novel designs in FERSNet. Specifically, we establish the FERSNet model without the FES regularization and without ConvFLU, respectively. We further employ the FES branch as a data augmentation approach for the FES task. The results are listed in Table \ref{tab:3}.

\begin{table}[ht]
    \centering
    \caption{The recognition results (\%) for ablation study. The best results are highlighted in \textbf{bold}.}
    \begin{tabular}{|l|c c c |}
    \hline
        Method & CK+ & Oulu-CASIA & MMI\\
        \hline
        FERSNet w/o MTL & 94.70 & 73.33 & 63.78 \\
        FERSNet w/o ConvFLU & 95.21 & 77.92 & 69.07 \\
        FERSNet (original) & 97.35 & 83.47 & 71.31 \\
        FERSNet w/ FES-DA & \textbf{97.75} & \textbf{87.64} & \textbf{73.87} \\
        \hline
    \end{tabular}
    \label{tab:3}
\end{table}
The FERSNet model, trained without FES regularization, becomes a single-task network, which does not adopt the multi-task learning strategy (FERSNet w/o MTL). The FERSNet model without ConvFLU (FERSNet w/o ConvFLU) becomes a hard parameter-sharing multi-task network, which does not obtain the ability to select beneficial information when transferring features. FERSNet with FES data augmentation (FERSNet w/ FES-DA) is the original FERSNet further fine-tuned on the synthetic samples produced by the FES branch. In other words, we employ the FES branch as a data augmentation approach for fine-tuning the network. As listed in Table \ref{tab:0}, MMI and CK+ are highly imbalanced datasets, and therefore we employ the FES branch to enlarge and balance the number of samples in each emotion class. For each training sample, we synthesize all the possible expression images from it to make the training set completely balanced. In Table \ref{tab:3}, we present the results of the ``FERSNet w/ FES-DA'' model using 24K images for training. It is worth noting that all the methods in Table \ref{tab:3} are based on the original FERSNet, without pre-training on BU-4DFE. To make a fair comparison, we enlarge the kernel size and network depth in ``FERSNet w/o MTL'' and ``FERSNet w/o ConvFLU'' to make their model capacity equal to or larger than the original FERSNet.

It is obvious from the table that jointly learning with FES contributes significantly to the FER performance, as the original FERSNet outperforms ``FERSNet w/o MTL'' by about 3\%, 10\%, and 8\% on CK+, Oulu-CASIA, and MMI, respectively. In addition, the proposed soft parameter-sharing strategy can further enhance the performance of MTL, because, without ConvFLU, the performance will be degraded by more than 2\% on the three datasets. More importantly, the FES branch can serve as a data augmentation method for FER. The original FERSNet, fine-tuned on the synthetic samples, acquires a better generalization ability for FER. With the FES data augmentation, the model can further obtain an accuracy gain of about 0.4\%, 4\%, and 2.5\% on CK+, Oulu-CASIA, and MMI, respectively.

\section{Conclusion}
In this paper, we have proposed a multi-task network, namely FERSNet, for facial expression recognition (FER) and facial expression synthesis (FES). FERSNet aims to solve the two tasks in parallel with the proposed convolutional feature leaky units (ConvFLU). ConvFLU adopts a soft parameter-sharing strategy, in order to filter out the useless and harmful features, when transferring information between FER and FES. Moreover, we further employ the FES branch for data augmentation to enlarge and balance the training dataset. This augmentation approach contributes to a better generalization of FERSNet for recognizing facial expressions in real-world applications. We evaluate the proposed method on three commonly used benchmarks. The experimental results have demonstrated that the proposed method achieves state-of-the-art performance, which makes it a potential solution to practical facial image analysis problems.

% references section

% can use a bibliography generated by BibTeX as a .bbl file
% BibTeX documentation can be easily obtained at:
% http://mirror.ctan.org/biblio/bibtex/contrib/doc/
% The IEEEtran BibTeX style support page is at:
% http://www.michaelshell.org/tex/ieeetran/bibtex/
\bibliographystyle{IEEEtran}
% argument is your BibTeX string definitions and bibliography database(s)
\bibliography{mybib.bib}
%
% <OR> manually copy in the resultant .bbl file
% set second argument of \begin to the number of references
% (used to reserve space for the reference number labels box)
% \begin{thebibliography}{1}

% \bibitem{IEEEhowto:kopka}
% H.~Kopka and P.~W. Daly, \emph{A Guide to \LaTeX}, 3rd~ed.\hskip 1em plus
%   0.5em minus 0.4em\relax Harlow, England: Addison-Wesley, 1999.

% \end{thebibliography}
\appendix
\begin{figure}[ht]
    \centering
    \includegraphics[angle =90, width=0.65\linewidth]{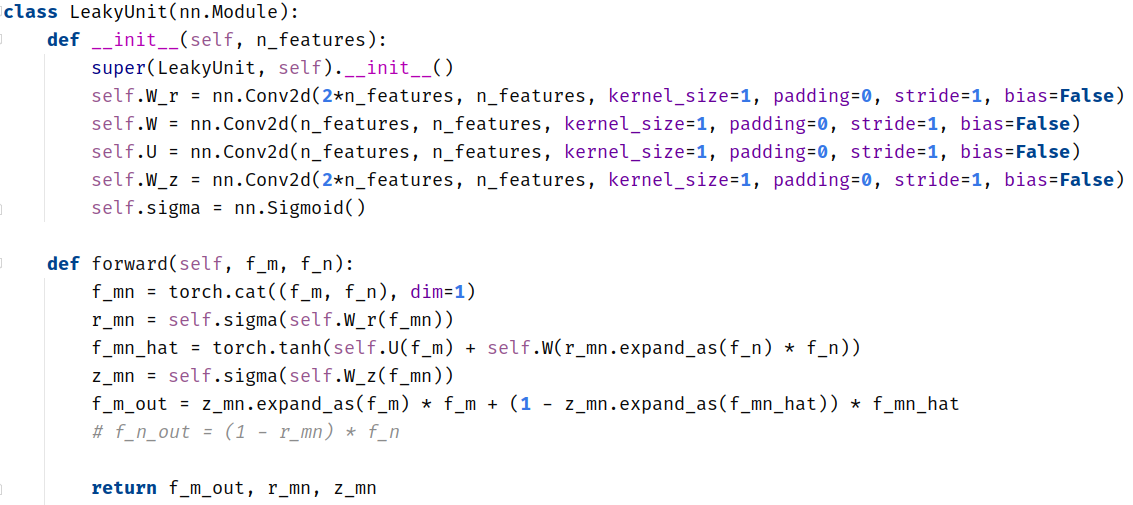}
\end{figure}

% that's all folks
\end{document}